\documentclass[runningheads]{llncs}
\usepackage{graphicx}
\usepackage{times}
\usepackage{booktabs}
\usepackage{latexsym}
\usepackage{amsmath}
\usepackage{url}
\usepackage[T5]{fontenc}
\usepackage[utf8]{inputenc}
\usepackage{multirow}
\usepackage{amsmath,amssymb,amsfonts}
\usepackage{algorithmic}
\usepackage{textcomp}
\usepackage[dvipsnames]{xcolor}
\usepackage{wrapfig,lipsum}
\usepackage{comment}
\usepackage{hyperref}
\usepackage{color}

\usepackage{microtype}

\makeatletter
\newcommand{\printfnsymbol}[1]{%
  \textsuperscript{\@fnsymbol{#1}}%
}
\makeatother

\begin{document}

\title{Automatic Textual Normalization for Hate Speech Detection}

\author{Anh Thi-Hoang Nguyen \inst{1,2,3,} \thanks{These authors contributed equally to this work}   \and
        Dung Ha Nguyen \inst{1,2,3,} \printfnsymbol{1}  \and
		Nguyet Thi Nguyen \inst{1,2,3}  \and
            Khanh Thanh-Duy Ho \inst{1,2,3}  \and
            Kiet Van Nguyen \inst{1,2,3} 
}

\authorrunning{Anh and Dung et al.}

\institute{Faculty of Information Science and Engineering, University of Information Technology, \\Ho Chi Minh City, Vietnam.  \and
Vietnam National University, Ho Chi Minh City, Vietnam. \and \email{20520134@gm.uit.edu.vn, 20520165@gm.uit.edu.vn, 20521689@gm.uit.edu.vn, 20521445@gm.uit.edu.vn, kietnv@uit.edu.vn}}

\maketitle 
\begin{abstract}
Social media data is a valuable resource for research, yet it contains a wide range of non-standard words (NSW). These irregularities hinder the effective operation of NLP tools. Current state-of-the-art methods for the Vietnamese language address this issue as a problem of lexical normalization, involving the creation of manual rules or the implementation of multi-staged deep learning frameworks, which necessitate extensive efforts to craft intricate rules. In contrast, our approach is straightforward, employing solely a sequence-to-sequence (Seq2Seq) model. In this research, we provide a dataset for textual normalization, comprising 2,181 human-annotated comments with an inter-annotator agreement of 0.9014. By leveraging the Seq2Seq model for textual normalization, our results reveal that the accuracy achieved falls slightly short of 70\%. Nevertheless, textual normalization enhances the accuracy of the Hate Speech Detection (HSD) task by approximately 2\%, demonstrating its potential to improve the performance of complex NLP tasks. Our dataset is accessible for research purposes \footnote{Github: \url{https://github.com/AnhHoang0529/Small-LexNormViHSD}}.

\end{abstract}

\section{Introduction}
\label{sec1}

The growing influx of social media users attracts researchers to conduct further studies, capitalizing on the abundant user-generated content. However, social media text often comprises non-standard language, abbreviations, slang, and emoticons, which can pose challenges in interpretation. Therefore, it is imperative to normalize social media text, which diminishes word variations, rendering it more accessible for NLP models to discern patterns, relationships, and meanings. In our research, lexical normalization is defined as the process of transforming text data into a more standardized or canonical form. This procedure aids in bridging the gap between the informal, dynamic nature of social media content and the necessity for structured and interpretable data for various NLP downstream tasks, including hate speech detection, sentiment analysis, machine translation, and more.

While textual normalization for social media has seen significant exploration in various languages, particularly in English, the literature review on Vietnamese is relatively limited. Most current systems designed for Vietnamese text normalization are customized for text-to-speech (TTS) applications and primarily utilize data extracted from newspapers or Wikipedia. It is essential to note that the definition of NSW in TTS differs from that in our study, as it encompasses elements such as email addresses, URLs, numbers, and more. Consequently, the methods mentioned in those studies are not directly applicable to our specific scenario. Furthermore, several studies have been conducted on social media, primarily involving data collected from Twitter, which is not a widely used social platform in Vietnam. Several studies often rely on manually crafted rules \cite{nguyen2020exploiting}, which may not encompass all Vietnamese NSW. In contrast, recent research frequently employ a two-step approach involving a detection stage and a normalization stage. Though this method handles the problem well, its architecture is too complex and domain-dependent. To the best of our knowledge, our study represents the initial endeavor to tackle Vietnamese lexical normalization within the social media context and to propose the implementation of a single Seq2Seq model for the given task.

Within the scope of our research, we aim to make three contributions:
\begin{itemize}
    \item We created a dataset that addresses the issue of normalizing Vietnamese social media text. The dataset includes pairs of comments in a non-standard and manually-labeled standard format. Table \ref{tbl:samples} presents an example of both non-standard and standard sentences.
    \item We employed Seq2Seq models to translate Vietnamese comments from a nonstandard structure to a standard form.
    \item We assume that textual normalization is generally enhance the performance of NLP downstream tasks. We validated this hypothesis by examining the Hate Speech Detection (HSD) task on datasets with and without lexical normalization.
\end{itemize}

\begin{table}[]
\label{tbl:samples}
\centering
\caption{A pair of non-standard and standard sentences.}
\begin{tabular}{ll}
\toprule
\textbf{Non-standard}     & mình cần tìm   gấp \textcolor{red}{in4} anh \textcolor{blue}{ng} \textcolor{ForestGreen}{iu} thất lạc ạ !                             \\
\midrule
\textbf{Standard} & mình cần tìm   gấp \textcolor{red}{thông tin} anh \textcolor{blue}{người} \textcolor{ForestGreen}{yêu} thất lạc ạ !                   \\
\midrule
\textbf{English}      & I need to seek information about my missing lover as soon as possible ! \\
\bottomrule
\end{tabular}%
\end{table}

\section{Related works}

The task of lexical normalization has been around since the early days of natural language processing, when researchers began working on text-to-speech and speech recognition systems. However, the challenges with traditional approaches, such as hand-crafted rules or lexicon-based methods, include word ambiguity and handling large datasets, though they can perform well in small datasets within specific domains. In the digital age, research on textual normalization in social media has inherited earlier achievements and developed more complex methodologies to address this task. In WNUT 2015 (Workshop on Noisy User-generated Text) \cite{bib6}, a shared task on Twitter Lexical Normalization for Named Entity Recognition was presented. Participants applied various approaches, including rule-based methods, machine learning techniques, and hybrid models that combined both approaches. In WNUT 2021 \cite{bib7}, the scope was expanded to include multilingual normalization and various downstream tasks. The results demonstrated that pre-trained models and encoder-decoder architectures are essential for achieving state-of-the-art results in lexical normalization.

Research in Vietnamese has also seen significant developments. Early studies focused on textual normalization using language models, hand-constructed dictionaries, and regular expressions \cite{normalize_vtweet,huongetal}. In recent years, researchers have exhibited a growing interest in deep learning models. The BiLSTM model served as the core component for correcting Vietnamese consonant misspellings \cite{10.1007/978-981-15-6168-9_40}. Transformer architecture was employed to address issues related to mistyped and misspelled errors \cite{do2021vsec}. Several publications in 2022 explored the utilization of the state-of-the-art BERT model and its combination with other architectural elements to create hybrid frameworks \cite{bertlm,ngo2022combination}. However, as mentioned in Section \ref{sec1}, none of the previous studies have considered the challenges posed by noisy social media data from common platforms in Vietnam today, such as Facebook or Youtube. Consequently, the techniques mentioned may not be suitable for our specific case study.

Drawing inspiration from the research \cite{bib4}, we treated lexical normalization as machine translation and implemented the Seq2Seq model to tackle the issue. This choice proves excellent for handling variations in user-generated content. The Seq2Seq model is also domain-independent, allowing it to process data for HSD in our case and to be adapted for data in other downstream tasks.

\section{Dataset Creation}

\subsection{Data Collection}
Social media text was extracted from the ViHSD dataset \cite{bib5}. This dataset contains 33,400 comments extracted from Facebook and Youtube sites and labeled as HATE, OFFENSIVE, or CLEAN. We specifically selected the first 3,400 comments from the training set of ViHSD\footnote{ViHSD dataset: \url{https://github.com/sonlam1102/vihsd}} to address the task of normalizing social media texts.

\subsection{Data Preprocessing}
Preprocessing the data is necessary to get it ready for labeling. Spaces are used to separate the syllables in the sentence. As an illustration, the sentence \textit{Hôm nay tôi đi chơi} (I am going out today) is converted into a collection of monosyllables \textit{['Hôm', 'nay', 'tôi', 'đi', 'chơi']}. Punctuation marks, brackets, emoji, emoticons, and other special characters are considered a single syllable and should be separated. However, in some cases, the string containing special symbols remains unchanged: emoticon, URL, email address, hashtags (\#), mention (@), time, date, and currency.

\subsection{Data Annotation}
Our labeling process and inter-annotator agreement (IAA) assessment are referenced from the studies \cite{bib6,bib7}. A group of 5 annotators labeled the dataset of 3,400 comments. All participants have completed a 12-year education, possess fluency in Vietnamese, and are well-acquainted with social network data. Each annotator labeled 1360 comments, and two people labeled each comment. To ensure autonomy in labeling, individuals utilize the Google Sheet tool to assign labels on separate files.

Typically, a comment is brief and does not provide enough context for the sentence. Hence, the process of normalizing text may vary significantly for each annotator, as it relies on individual proficiency in Vietnamese, writing habits, and sensitivity to recent news. In order to maximize the IAA, we established guidelines for data annotation, which are available for reference \footnote{Guideline:  \url{https://bit.ly/TextNormAnnotationGuideline}.}.

The annotation process took approximately two weeks to finish. We used two measures in order to evaluate IAA. \textbf{(1)}\textit{ The Cohen's Kappa coefficient} is employed to determine the agreement regarding whether to normalize that specific token. \textbf{(2)} \textit{Same candidates} quantifies the frequency at which annotators align in their textual normalization. The value of \textit{Cohen's Kappa coefficient} is 0.9014, revealing a high level of IAA in textual normalization. \textit{Same candidates} was measured at 0.7449, which implies a high level of agreement in token modification between two annotators. The outcome indicates that the normalization process is generally and uniformly effective, and the text has been normalized consistently and accurately.

In several cases, tokens fail to reach a consensus on their standard form. If one annotator shows certainty in the normalization while another person does not, the former assigns the final normalized token. In the next stage, a different person, who was not involved in the previous labeling process, re-labeled the remaining cases. Finally, we withdrew any comments with no non-standard tokens from the dataset and duplicated sentences with multiple modifications, changing the dataset's size from 3,400 to 2,181 comments.

\subsection{Error Analysis of Annotation}

During the labeling process, annotators made multiple mistakes due to carelessness, non-compliance with guidelines, and the context's ambiguity. We provide several errors made by the annotators for illustration in Table~\ref{annotationerrors}

\begin{table}[h]
\centering
\caption{Annotation Errors.}
\label{annotationerrors}
\resizebox{\columnwidth}{!}{%
\begin{tabular}{ccccl}
\hline
\textbf{sentence}   & \textbf{token} & \textbf{norm} & \textbf{is\_norm} & \multicolumn{1}{c}{\textbf{annotation errors}}                             \\ \hline
\multirow{3}{*}{20}  & bamtp & bạn    & \# & Ambiguity: unclear sentence                                                                  \\
                     & chạy  & chạy   &    &                                                                                              \\
                     & ddi   & đi     &    & Non-compliance with guidelines: '\#' should be in column 'is\_norm', following the guideline \\ \hline
\multirow{3}{*}{65} & abe            & anh bé        & \#                & Ambiguity: 'abe' should remain unchanged as this is the name of a streamer \\
                     & 20-10 & 20-11  &    & Carelessness: unintentionally change the standard token, '20-10' should remain unchanged     \\
                     & vv    & vui vẻ & \# &                                                                                              \\ \hline
\multirow{4}{*}{110} & cận   & cẩn    & \# &                                                                                              \\
                     & thận  & thận   &    &                                                                                              \\
                     & nhà   & nhà    &    & Carelessness: miss a NSW, 'nhà' should be modified as 'nha'                                  \\
                     & ae    & anh em & \# &                                                                                              \\ \hline
\end{tabular}%
}
\end{table}



\subsection{Dataset Statistics}
The dataset consists of 2,181 pairs of original and normalized comments and was randomly divided into train, validation, and test sets with the ratio of 8:1:1. Table~\ref{tab1} shows the total number of tokens and non-standard tokens that need to be normalized in the data set.

\begin{table}[h]
\centering

\caption{Dataset Statistics.}
\begin{tabular}{lcccc}
\hline
\textbf{}                                                                                         & \multicolumn{1}{c}{\textbf{Train}} & \multicolumn{1}{c}{\textbf{Valid}} & \multicolumn{1}{c}{\textbf{Test}} & \multicolumn{1}{c}{\textbf{Full Data}} \\ \hline
\textbf{Number of tokens}                                                                         & 26,990                              & 3429                               & 3,569                              & 33,988                                  \\
\textbf{Number of tokens need to be normalized}        & 4,901                               & 589                                & 641                               & 6,131                                   \\
\textbf{Number of unique tokens}                                                                  & 3,704                               & 1,208                               & 1,259                              & 4,103                                   \\
\textbf{Number of unique tokens need to be normalized} & 1,452                               & 290                                & 329                               & 1,638                                   \\ \hline
\end{tabular}%

\label{tab1}
\end{table}

The most frequent tokens that require normalization in the data set are shown in Table~\ref{tab2}. The word 'ko' (not) occurs the most, 235 times, accounting for 3.8\% in the training set, 2.7\% in the validation set, and 4.8\% in the test set.

\begin{table}
\centering
\caption{Most Common Non-standard Tokens in the Dataset.}
\begin{tabular}{ccccccccc}
\hline
\textbf{Rank} &
  \multicolumn{2}{c}{{ \textbf{Train}}} &
  \multicolumn{2}{c}{\textbf{Valid}} &
  \multicolumn{2}{c}{\textbf{Test}} &
  \multicolumn{2}{c}{\textbf{Full Data}} \\ \hline
1  & ko & 188 & t  & 27 & t  & 38 & ko & 235 \\
2  & dm & 168 & k  & 22 & ko & 31 & t  & 231 \\
3  & t  & 166 & dm & 18 & đc & 26 & dm & 205 \\
4  & đc & 128 & m  & 18 & m  & 25 & đc & 171 \\
5  & k  & 120 & đc & 17 & dm & 19 & k  & 160 \\
6  & vl & 93  & ko & 16 & k  & 18 & m  & 133 \\
7  & m  & 90  & e  & 10 & a  & 15 & vl & 112 \\
8  & a  & 78  & vl & 10 & vn & 12 & a  & 101 \\
9  & e  & 72  & đm & 10 & e  & 9  & e  & 91  \\
10 & vn & 58  & vn & 8  & vl & 9  & vn & 78  \\ \hline
\end{tabular}%
\label{tab2}
\end{table}
\section{Methodology}
\subsection{Normalization methods}
In this study, we defined the lexical normalization task as a machine translation problem involving translating a non-standard sentence into a standard one. We applied the word-level Seq2Seq model (S2S) and its two variants S2SSelf and S2SMulti, introduced in \cite{bib4}. Since the model works at the token level, the monosyllables are treated as the smallest unit in the input, target, and output sequences. A non-standard comment is represented as an input sequence of $T$ syllables $\vec{x} = [x_1, x_2,..., x_T]$, Seq2Seq models generate an output sequence $\vec{y} = [y_1, y_2,..., y_L] $ having length $L$ with the same meaning as $\vec{x}$.

\subsubsection{S2S model:}
The architecture of the S2S model is based on the encoder-decoder framework \cite{bib8,bib9}. We utilized the Bi-LSTM model with the attention mechanism added to normalize social media text.

The encoder takes an input string $\vec{x}$ and turns it into a contextual hidden state string $\vec{h} = [h_1, h_2,..., h_T]$. The bidirectional model employs two encoders, each of which processes the text forward and backward. The ultimate hidden state at time step $t$ is a combination of two encoder components $\vec{h_t}=[g_f\left(x_t,\ h_{t-1}\right);\ g_b\left(x_t ,\ h_{t+1}\right)]$, where $g_f$ and $g_b$ are forward and backward encoder units. Similar to how the encoder does, the decoder also defines a series of hidden states $\vec{s_j}=g_s(s_{j-1},{\ y}_{j-1},\ c_j)$, where ${\ y}_{j-1}$ is the preceding token, ${\ s}_{j-1}$ is the decoder state and $c_j$ is the context vector, which is computed by the weighted sum of the hidden states leveraging the attention mechanism \cite{bib10}:

\begin{equation}
c_j = \ \sum_{i=k}^{\mid t \mid}{\alpha_{jk}h_k}\label{eq1}
\end{equation}

where $\alpha_{jk}=Softmax(f(s_{j-1},\ h_k))$ and $f\left(s_{j-1},\ h_k\right)=s_{j-1} ^TWh_k$ is a generalized function according to \cite{bib10}. Then each output syllable is predicted by the Softmax classifier $y_j\ \sim \ p\left(y_j\mid y_{<j},\ \vec{x}\right)=Softmax(\psi \left(s_j\right))$, where $\psi$ is an affinity function that transfers the decoder state to a vector of vocabulary size.

\begin{equation}
L\left(\theta\right)=\ -\sum_{\left(\vec{x},\ \vec{y}\right)\in D}\sum_{j=1}^{\mid L \mid}{logp_\theta}(y_j\mid y_{<j},\ \vec{x})\label{eq2}
\end{equation}

A bad prediction made during training can result in compounded mistakes in following training steps. So, when calculating the conditional probability $p_\theta\left(y_j\mid y_{<j},\ \vec{x}\right)$, it is common to use the Scheduled Sampling method \cite{bib11}. This strategy alternates between using the model's prediction results in the previous step ${\hat{y}}_{j-1}$ and its corresponding output $y_{j-1} $ to reduce accumulated errors.

The token-level Seq2Seq model can capture token-level semantic content and persistent context dependencies that accurately map words with different modifications. Depending on the circumstances, the relevant normalized input, target, and output sequences based on the context are shown in Table~\ref{fig4}.

\begin{table}[h]
\centering
\caption{Illustration of an Input Sequence, Target Sequence, and Output Sequence.}
\begin{tabular}{lllllllllllll} 
\hline
\textbf{Input}  & mình & cần & tìm & gấp & \textcolor[rgb]{1,0.647,0}{in4}       & anh & \textcolor[rgb]{1,0.647,0}{ng}    & \textcolor[rgb]{1,0.647,0}{iu}  & thất & lạc & ạ & !  \\ 
\hline
\textbf{Target} & mình & cần & tìm & gấp & \textcolor[rgb]{0,0.502,0}{thông tin} & anh & \textcolor[rgb]{0,0.502,0}{người} & \textcolor[rgb]{0,0.502,0}{yêu} & thất & lạc & ạ & !  \\ 
\hline
\textbf{Output} & mình & cần & tìm & gấp & \textcolor{red}{in4}       & anh & \textcolor[rgb]{0,0.502,0}{người} & \textcolor[rgb]{0,0.502,0}{yêu} & thất & lạc & ạ & !  \\ 
\hline
                &                                    &                                   &                                   &                                   &                                         &     &                                     &                                   &                                    &                                   &                                 &   
\end{tabular}%
\label{fig4}
\end{table}

\subsubsection{S2SSelf model:}
A special symbol, '@self' marks the token in the input string unchanged during training. For example, the standard tokens in the input string 'Nc là chất điện phân' are replaced with '@self', so the input string will become 'Nc @self @self @self @self'. We replace this special symbol with the corresponding word in the original sentence when making predictions.
\subsubsection{S2SMulti model:}
To begin, we convert tokens with a single normalized format and then translate tokens with many adjustments using the S2S model.

\subsection{Experimental Settings}

We experimented with three models using the original implementation\footnote{Github: \url{https://github.com/Isminoula/TextNormSeq2Seq}} with several changes in parameter settings. The 'val\_split' was set to 219, and 'max\_train\_decode\_len' was set to 165 in all experiments. Parameter 'end\_epoch' of S2SSelf and S2SMulti were set to 100 and 80, respectively.

\subsection{Evaluation Metrics}

\textbf{F1 score} is employed to assess the appropriate labeling rate for tokens that need to be normalized. We calculated the F1 score on each pair of sentences and took the average value. For sentences with multiple modifications, that is, pairs of sentences with the same input and different targets, we took the highest F1 value.

Precision $p = \textit{correct norm}/\textit{total norm}$ and recall $r = \textit{correct norm}/\textit{total nsw}$, where \textit{correct norm} is the number of tokens that are appropriately normalized, \textit{total norm} is the number of normalized tokens in the prediction sentence, i.e., there is a change from the original sentence, and \textit{total nsw} is the total number of tokens that need to be normalized in the input string. F1 score is calculated based on precision and recall $f1 = (2 \times p \times r)/(p + r)$.

\textbf{BLEU-4} is a suitable algorithm for evaluating the performance of a machine translation task, as introduced in \cite{papineni-etal-2002-bleu}. We leveraged the BLEU-4 score as another metric to compare the results of baseline models.

\subsection{Experimental Results}
Table~\ref{tab3} shows that the textual normalization results are not impressive since all metrics are lower than 70\%. The S2SMulti model has the best ability in textual normalization, as its F1-score and BLEU-4 on the validation and test sets are the highest. The results obtained on the S2S model are close to the S2SMulti model. However, the S2SSelf model received the lowest results on both metrics, about 40\% for the F1-score and less than 60\% for the BLEU-4.

The input string lacks important information when substituting \@self into standard tokens. Therefore, S2SSelf cannot take advantage of the context and misinterprets the comments. S2SMulti ensures that tokens with a unique interpretation are correctly transformed during learning, increasing the sentence context's reliability in the learning process. Nevertheless, the traditional S2S model does not guarantee the correct translation of tokens with one interpretation. As a result, in some cases, S2SMulti provides better translations.
\begin{table}[h]
\centering
\caption{Experimental Results (\%) of Lexical Normalizaion.}\label{tab3}
\begin{tabular}{lcccccc}
\hline
\multirow{2}{*}{} & \multicolumn{2}{c}{\textbf{S2S}} & \multicolumn{2}{c}{\textbf{S2SSelf}} & \multicolumn{2}{c}{\textbf{S2SMulti}} \\ \cline{2-7} 
                  & F1              & BLEU-4         & F1                & BLEU-4           & F1                & BLEU-4            \\ \hline
Valid             & 54.26          & 65.04         & 40.88            & 57.56           & 55.63            & 65.26            \\
Test              & 56.16          & 68.71         & 37.65            & 58.40           & 58.30            & 69.71            \\ \hline
\end{tabular}%
\end{table}

\subsection{Error Analysis of Normalization}
\begin{figure}[h]
    \centering
    \caption{Effect of sentence length on the error rate of models.}
    \includegraphics[width=0.9\linewidth]{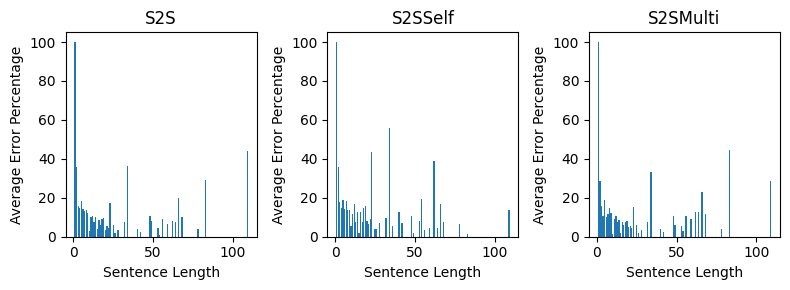}
    \label{fig5}
\end{figure}

From Figure~\ref{fig5}, we find that if the sentence is too long or too short, the model's error rate is higher. In this case, the ideal length of the strings fed into the model is between 25 and 50. Given that sequence length significantly impacts the error rates of models, sentence length treatment can be an essential aspect to improve performance and reduce the error rates of these models. 

Some labeling errors can affect the model's prediction results. The most frequent case is asynchronous labeling for sentences with multiple normalizations. For example, the token 't' (I) can be interpreted as 'tôi', 'tao', or 'tớ'. A sequence in the testing set 'chỉ t vs' (help me) was labeled with only one version: 'chỉ tao với'. The model's output, however, was 'chỉ tôi với', which has a proper meaning but does not match the target. Therefore, the accuracy in this case is 0 despite the correct translation.

\subsection{Leveraging Automatic Textual Normalization for Hate Speech Detection}

\begin{figure}[h]
    \centering
    \caption{Automatic textual normalization for Hate Speech Detection.}
    \includegraphics[width=\linewidth]{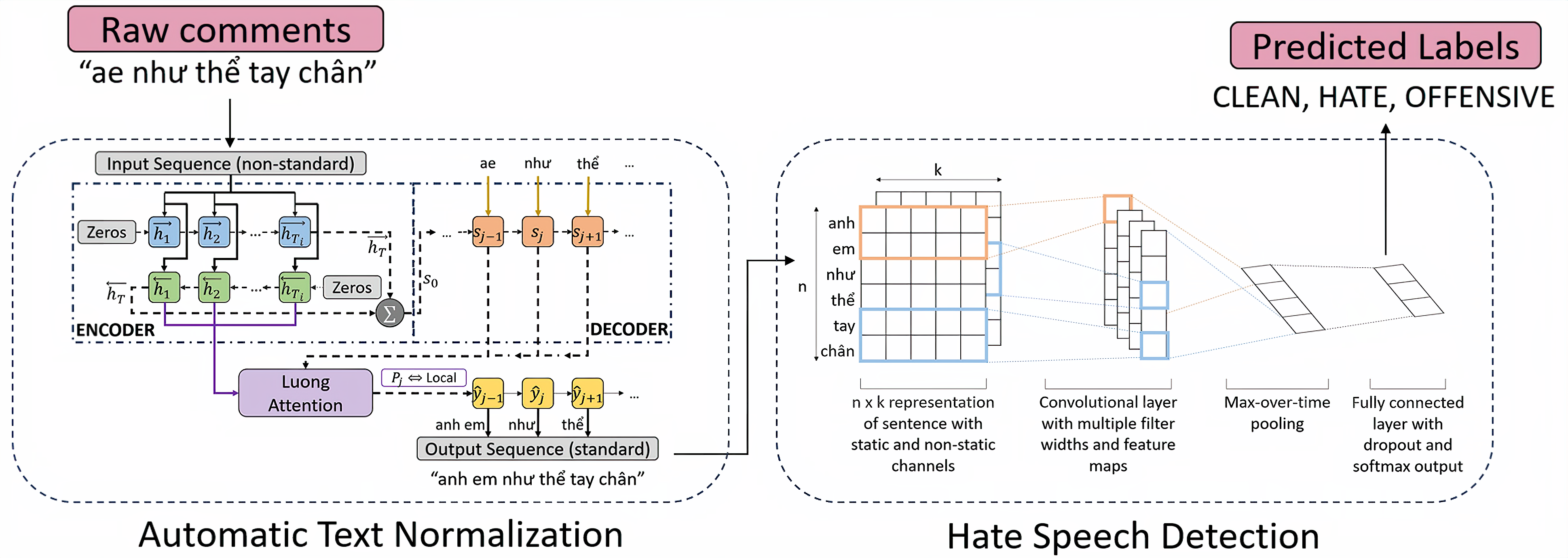}
    \label{fig6}
\end{figure}

This experiment aims to assess the importance of textual normalization as a preprocessing stage for additional challenging NLP tasks, namely hate speech detection (HSD). The challenge with HSD is detecting harmful content on social media. The learning model receives the comments, processes them, and outputs the labels, including CLEAN, OFFENSIVE, and HATE. We predicted labels using Text-CNN and GRU and evaluated the performance using F1-micro, F1-macro, and Accuracy, as used in the original paper. \cite{bib5}. We also ensured that all parameters matched precisely those in the original research.

The experiment was performed on unnormalized and normalized comments, respectively. We obtained normalized data from the results of the S2SMulti model, which had the best result. The experimental dataset contains 438 standard comments combined from the validation (219 sentences) and the test set (219 sentences). The non-standard comments were collected corresponding to the above 438 standard comments. We extracted the labels corresponding to each comment from the ViHSD dataset. Finally, the ratio of training, validation, and test sets of both experiments was 8:1:1.

Table~\ref{tab4} shows that using textual normalization improves the accuracy of HSD tasks by 2\% on F1-micro and Accuracy, about 6 - 7\% on F1-macros for both Text-CNN and GRU models. We concluded that compared with the usual simple preprocessing steps, textual normalization increases HSD performance in this case. Although the accuracy of textual normalization is not very high (less than 70\%), this experiment shows the potential of textual normalization in supporting other complex NLP tasks.

\begin{table}[h]
\centering
\caption{Experimental Results (\%) of Hate Speech Detection (HSD).}
\label{tab4}
\resizebox{0.9\textwidth}{!}{%
\begin{tabular}{ccccccc}
\hline
\textbf{} & \multicolumn{3}{c}{{ \textbf{Text-CNN}}}           & \multicolumn{3}{c}{\textbf{GRU}}                                       \\ \hline
          & \textbf{Unnormalized} & \textbf{Normalized} & \textbf{Improvement} & \textbf{Unnormalized} & \textbf{Normalized} & \textbf{Improvement} \\ \hline
\textbf{F1-micro} & 77.27 & 79.55 & \color{blue}{+2.28} & 75.00   & 77.27 & \color{blue}{+2.27} \\
\textbf{F1-macro} & 35.61 & 41.45 & \color{blue}{+5.84} & 28.57 & 35.61 & \color{blue}{+7.04} \\
\textbf{Accuracy} & 77.27 & 79.55 & \color{blue}{+2.28} & 75.00   & 77.27 & \color{blue}{+2.27} \\ \hline
\end{tabular}%
}
\end{table}

\section{Conclusion and Future Direction}
In this study, we created a dataset and models to address the issue of normalizing social media Vietnamese text. In particular, we gained a good IAA in the annotation phase, which ensures the quality of the dataset for deep learning models. Furthermore, we normalized non-standard tokens by using sequence-to-sequence models. The findings demonstrate that the machine models were moderately successful but challenging in normalizing text as all metrics are below 70\%. However, our research shows it sufficiently boosted the HSD's performance.

However, the dataset size is relatively small, which may cause a limitation in lexical normalization. Manually labeling the data was laborious and time-consuming. Therefore, in future works, we aim to enlarge the dataset with a semi-supervised approach to save time and effort. Another challenge in our study is the long training time of Seq2Seq models, so we concentrate on exploiting other deep learning methods to solve the given task effectively. We also wish to broaden the scope of this study to other NLP tasks besides HSD, which strengthens the supporting role of lexical normalization in multiple NLP applications.

\emergencystretch=2em


\bibliographystyle{splncs04}
\bibliography{bibliography}

\end{document}